\documentclass[a4paper]{article}

\usepackage{INTERSPEECH2022}
\usepackage{multirow}
\usepackage{color}
\usepackage[hyphens]{url}
\usepackage[hidelinks]{hyperref}
\usepackage[hyphenbreaks]{breakurl}
\usepackage[capitalize]{cleveref}
\usepackage{enumitem}
\usepackage{etoolbox}
\usepackage{graphicx}
\usepackage{makecell}
\usepackage{caption}
\usepackage{wrapfig}
\usepackage{setspace}
\usepackage{subcaption}
\usepackage{floatflt}
\usepackage{float}
\usepackage{tikz}
\usepackage{pgfplots}
\usetikzlibrary{positioning,chains,shapes.geometric, calc, shadows, shapes.misc}
\usepackage[super]{nth}

\newcommand{\concat}{Concat}
\newcommand{\simpleAdd}{Simple-Add}
\newcommand{\complexAdd}{Complex-Add}
\newcommand{\gatedAdd}{Gated-Add}
\newcommand{\weightedSimpleAdd}{Weighted-Simple-Add}

\title{Improving the Training Recipe for a Robust Conformer-based Hybrid Model}
\name{
  Mohammad Zeineldeen$^{1,2,*}$\thanks{$^*$Equal contribution},
  Jingjing Xu$^{1,*}$,
  Christoph Lüscher$^{1,2}$,
  Ralf Schlüter$^{1,2}$,
  Hermann Ney$^{1,2}$
}
\address{
  $^1$Human Language Technology and Pattern Recognition, Computer Science Department, \\
  RWTH Aachen University, 52074 Aachen, Germany \\
  $^2$AppTek GmbH, 52062 Aachen, Germany
}

\email{
  \{zeineldeen, luescher, schlueter, ney\}@cs.rwth-aachen.de,
  jingjing.xu@rwth-aachen.de
}

\renewrobustcmd{\bfseries}{\fontseries{b}\selectfont}
\renewrobustcmd{\boldmath}{}
\newrobustcmd{\B}{\bfseries}

\newsavebox\CBox

\makeatletter

\renewcommand{\section}{\@startsection
   {section}%
   {1}%
   {}%
   {-0.4\baselineskip}%
   {0.2\baselineskip}%
   {}}%

\renewcommand{\subsection}{\@startsection
  {subsection}%
  {2}%
  {}%
  {-0.1\baselineskip}%
  {0.1\baselineskip}%
  {}}%

\renewcommand{\subsubsection}{\@startsection
  {subsubsection}%
  {3}%
  {}%
  {-0.2\baselineskip}%
  {0.2\baselineskip}%
  {}}%

\begin{document}

\maketitle
\begin{abstract}
Speaker adaptation is important to build robust automatic speech recognition
(ASR) systems.
In this work, we investigate various methods for speaker adaptive training
(SAT) based on feature-space approaches for a conformer-based acoustic model
(AM) on the Switchboard 300h dataset.
We propose a method, called {\weightedSimpleAdd}, which adds weighted
speaker information vectors to the input of the multi-head self-attention
module of the conformer AM.
Using this method for SAT, we achieve $3.5\%$ and $4.5\%$ relative
improvement in terms of WER on the CallHome part of Hub5'00 and Hub5'01
respectively.
Moreover, we build on top of our previous work where we proposed
a novel and competitive training recipe for a conformer-based hybrid AM.
We extend and improve this recipe where we
achieve $11\%$ relative improvement in terms of word-error-rate (WER) on
Switchboard 300h Hub5'00 dataset.
We also make this recipe efficient by reducing the total number
of parameters by $34\%$ relative.
\end{abstract}

\section{Introduction \& Related Work}
% * 14% relative improvement: 12.5 -> 10.7
%
% * efficient training: reduced number of parameters from 88M to 53M which
% is 40% relative
%
% * we build a recipe including LN instead of BN which fits better for streaming
% and also not depending on batch size
%
% * we propose a new method to integrate i-vectors speaker features into
% the conformer system which yields better results
%
% * we investigate AT stuff
%
% * we do analysis between BLSTM and Conformer regarding speaker error

Hybrid neural network (NN)-Hidden Markov model (HMM) has been
widely used to build competitive automatic speech recognition (ASR) systems
on different datasets
\cite{zeineldeen2022:hybrid-conformer,zhou2020rwth,Lscher2019RWTHAS}.
The system consists of an acoustic model (AM) and a language model (LM).
Due to the tremendous success of deep learning, NN architectures
are used for acoustic modeling.
Such NN architectures include Bidirectional long short-term
long memory (BLSTM) \cite{hochreiter1997long},
convolution neural networks (CNN) \cite{abdel2014convolutional},
and time-delay neural networks (TDNN) \cite{peddinti2015time}.
The NN acoustic models are often trained with cross-entropy using a
frame-wise alignment generated by a Gaussian mixture model (GMM)-HMM
system.

% extention to our recipe
Recently, self-attention models \cite{Sperber2018SelfAttentionalAM},
specifically, conformer-based models \cite{gulati2020conformer},
have achieved state-of-the-art performance on different
datasets \cite{gulati2020conformer,Zoltan2021conformer}.
In \cite{zeineldeen2022:hybrid-conformer}, we proposed a novel and
competitive training recipe for building conformer-based hybrid-NN-HMM systems.
We studied different training methods to improve performance as well as
to speed up training time.
In this work, we extend and improve our training recipe by making it more
efficient in terms of training speed and memory consumption as well as having a
significant improvement in terms of word-error-rate (WER).

% proposed ways to do SAT
% issue: strong model do not benefit too much
% reference: zoltan's paper, sota result but does not benefit from i-vectors
Moreover, speaker adaptation methods have been used to build robust
ASR systems \cite{Saon2013SAT}.
These methods can be classified as feature-space approaches and
model-space approaches.
Feature-space approaches include using speaker adaptive features \cite{mllr} or
augmenting speaker information such as
i-vectors \cite{Dehak2011Ivector} or x-vectors \cite{snyder2018xvector} into
the network.
On the other hand, the model-space approaches modify the acoustic model to
match the testing conditions by learning a speaker or environment dependent
linear transformation to transform the input, output, or hidden representations
of the network \cite{seide2011feature, gemello2006adaptation, li2010comparison}.
% Feature-space approaches has some advantages as they use a unified trained
% model whereas model-space approaches may require a model per speaker which
% in some cases is not feasible.
There have been some attempts to apply feature-space approaches to
state-of-the-art conformer-based models \cite{Zoltan2021conformer}.
In \cite{Zoltan2021conformer}, they concatenate i-vectors to the input
of each feed-forward layer in the conformer block, however, this gives only
minor improvements and adds many additional parameters.
In addition, there has been other work related to multi-accent adaptation
which studied also different ways to integrate accent information into
the network \cite{Zhu2019multiaccent, gong2021multiaccent, deng2021improving,
turan2020multiaccent}.
However, in all these approaches, the conformer architecture was not used.
Therefore, in this work, we focus on feature-space
approaches and we investigate better ways to integrate speaker information
into the conformer AM.

% Summary of contributions
The main contributions of this paper are:
(1) We improve our training recipe for conformer-based hybrid-HMM-NN models
\cite{zeineldeen2022:hybrid-conformer} which lead to $11\%$ relative
improvement in terms of WER on Switchboard 300h Hub5'00 dataset,
(2) We build an efficient conformer AM with $34\%$ relative reduction
of the total number of parameters compared to our previous baseline
\cite{zeineldeen2022:hybrid-conformer},
(3) We propose a better method to integrate i-vectors into the conformer AM
and achieved $3.5\%$ relative improvement in terms of WER on CallHome part
of Switchboard 300h.
We call this method {\weightedSimpleAdd} and it adds weighted speaker
information vectors to the input of the multi-head self-attention module
of the conformer AM.

\section{Model Architecture}

\begin{figure}[t]
  \centering
  \includegraphics[width=\linewidth]{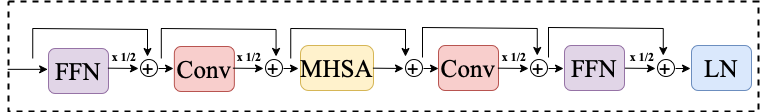}
  \caption{\textit{Overview of the proposed conformer block.}}
  \label{fig:conf-block}
\vspace{-6mm}
\end{figure}

Our AM is a variant of the conformer architecture \cite{gulati2020conformer}.
The training recipe is built on top of our previous work
\cite{zeineldeen2022:hybrid-conformer}.
The main differences are: (1) we replace all batch normalization (BN)
\cite{Ioffe2015bn} layers with layer normalization (LN) \cite{Lei2016ln} layers,
(2) we add another convolution module following Macaron
style \cite{gulati2020conformer}.
Each conformer block consists of one or more of these modules:
feed-forward (FFN) module, multi-head self-attention (MHSA) module, and
convolution (Conv) module.
The proposed conformer block is illustrated in \Cref{fig:conf-block}.
More details and experimental results are discussed in \Cref{sec:exp-results}.

\section{Speaker Adaptation}
\label{sec:sat-methods}

The main goal of speaker adaptation is to reduce the mismatch between train
and test data by adapting the model to the unseen speakers of the test data.
In this way, these methods adapt the speaker-independent model to be more
robust to different speakers observed in the test data.
In this work, we focus on feature-space adaptation methods.

Feature-space methods adapt the features of the network to be speaker-dependent.
I-vectors \cite{Dehak2011Ivector} have been successfully applied to NN-based
ASR systems \cite{Saon2013SAT,Peddinti2015tdnn}.
However, the integration of such vectors depends on the model
architecture itself and it is not always clear what is the best way to
do such an integration.
For the BLSTM model, it was observed that concatenating the i-vectors to the
network features gives significant improvement \cite{kitza2019cumulative}.
But this did not work well for the conformer model as observed in
\cite{Zoltan2021conformer}.
In this work, we investigate and propose better ways to integrate the
i-vectors into the conformer AM.
We develop a method to combine i-vectors into the MHSA module of the conformer
AM which gives the best improvement.

% Let $x_1^T$ be the input feature vectors for some utterance of length $T$
% frames.
Let $z_1^{T}$ be the input to a module in some conformer block
(more details can be found in \Cref{sec:sat-results}),
and let $v$ represents a speaker representation vector such
as i-vector or x-vector for the input utterance.
The methods to integrate $v$ into the network can be defined as:
\begin{itemize}

  % \item \featureConcat: concatenate the input feature vectors with i-vector $v$
  %   \begin{align*}
  %     \tilde{x}_t = [x_t; v] \quad \forall t \in [1, T]
  %   \end{align*}

  \item {\concat}: concatenate the hidden representation $z_t$ with $v$
  \begin{align}
    \tilde{z}_t = [z_t; v] \quad \forall t \in [1, T]
  \end{align}

  \item {\simpleAdd}: $U$ and $b$ are trainable parameters.
  \begin{align}
    \tilde{z}_t = z_t + U v + b \quad \forall t \in [1, T]
  \end{align}

  \item {\complexAdd}: $W$, $U$, and $b$ are trainable parameters.
  \begin{align}
    \tilde{z}_t = Wz_t + U v + b  \quad \forall t \in [1, T]
  \end{align}

  \item {\gatedAdd}: learn a scale and shift parameters conditioned on
  the speaker information $v$ \cite{yoo2019film, Rownicka2019}.
  % Then, these parameters are used to scale and shift $h_t$.
  \begin{align}
    \tilde{z}_t & = z_t \otimes \gamma + \beta  \quad \forall t \in [1, T] \\
    \gamma      & = \text{tanh}(W v) + b_1 \\
    \beta       & = \text{tanh}(U v) + b_2
  \end{align}
  where $W, U, b_1, b_2$ are trainable parameters.

  \item {\weightedSimpleAdd}: The idea is to compute a score between each
  hidden representation $z_t$ and the speaker representation vector $v$.
  Sigmoid function is used to get a score between 0 and 1.
  Inspired from \cite{deng2021improving}, we also apply a
  threshold $k$ on the learned weights.
  We use a value of 0.4 for $k$.
  \begin{align}
    w_t &= \sigma(z_t \cdot (\text{tanh}(W\cdot v) + b_1)) \in \mathbb{R} \\
    w_t &=
      \begin{cases}
        w_t, & w_t \geq k \\
        0,   & w_t < k
      \end{cases} \\
    \tilde{z}_t &= z_t + w_t \odot (U v + b_2) \quad \forall t \in [1, T]
  \end{align}
  where $\sigma$ is the sigmoid function.
  $\cdot$ and $\odot$ are the dot-product and element-wise product operations
  respectively.
  $W$, $U$, $b_1$, and $b_2$ are trainable parameters.
\end{itemize}

Note that the representation capability (learnable space of
parameters) is almost similar in nature between methods \concat, \simpleAdd,
and {\complexAdd} (see \Cref{sec:sat-results}).

% \subsection{Model-space Methods}
% Model space adaptation aims at adjusting the model parameters to alleviate the
% mismatch betwen train and test dataset.
% In our work, we choose to use affine transformation based methods which insert
% an extra linear layer with bias between any two layers in the source model.
% Given a pretrained acoustic model and an adaption dataset, the adaption set is
% first divided into different subsets based on the adaption label of the
% utterance.
% We use one affine transformation for each adaption label and train it with the
% corresponding data subsets.
% It could be formulated as:
% \[\hat{h} = W_a \cdot h + b_a\]
% where h represents internal features from the hidden layer in the acoustic model,
% $W_a$ and $b_a$ are the adaption label related weights and bias.
% From our experiments, we observed that using activations would hurt the
% performancem, so no activation is used here.
% We apply the affine transformation adaption on the acoustic model augmented
% with i-vector.
% +Model-space speaker adaptation methods aims at adapting parts of the model
% +to alleviate the mismatch betwen train and test dataset.
% +In this paper, we focus on adapting an inserted affine transformation
% +between the layers which is also known as Linear Hidden Network (LHN)
% +\cite{}.

\section{Experimental Setup}
Experiments are conducted on the Switchboard 300h dataset \cite{Godfrey1992SWB}
which consists of English telephony conversations.
We use Hub5'00 as a development set for tuning the search hyperparameters.
It consists of both Switchboard (SWB) and CallHome (CH) parts.
CH part is much noisier.
We use Hub5'01 as test set.
RETURNN \cite{Zeyer_2018} is used to train the acoustic models and
RASR \cite{Wiesler2014RASR} is used for decoding.
All our configs files and code to reproduce our results can be found
online\footnote{\scriptsize\url{https://github.com/rwth-i6/returnn-experiments/tree/master/2022-swb-conformer-hybrid-sat}}.

\subsection{Baseline}

We use the baseline from our previous
work \cite{zeineldeen2022:hybrid-conformer}.
It has 12 conformer blocks and we do time down-sampling by a factor of 3.
Up-sampling is applied to the output layer using transposed convolution
\cite{long2015fully} to match the target fixed alignment.
The total number of parameters is 88M.
More details regarding hyperparameters and training methods can be found
in \cite{zeineldeen2022:hybrid-conformer}.
We noticed that the momentum and epsilon parameters used in the BN layer
were not optimal.
After updating these parameters, our baseline achieves now WER of $11.7\%$ on
Hub5'00.
The details for applying speaker adaptation are explained in
\Cref{sec:sat-results}.

\subsection{I-vectors Extraction}

Our i-vector extraction pipeline follows the recipe described
in \cite{kitza2019cumulative}.
To train the universal background model (UBM), 40-dimensional Gammatone
features \cite{schluter2007Gammatone} with a context of 9 frames are
concatenated and then reduced to a dimension of 60 using
linear discriminant analysis (LDA).
Normalization and LDA are usually applied after the extraction.
We do not scale the i-vectors as done in \cite{kitza2019cumulative}
since we found this is crucial to get improvement.
The same observation is also stated in \cite{Rouhe2020}.
For the final LDA dimension reduction, we have compared using i-vectors
with dimensions 100, 200, and 300.
Using 200-dimensional i-vectors is the best for our settings.

\subsection{Sequence Discriminative Training (seq. train)}

We use the lattice-based version of state-level minimum Bayes risk (sMBR)
criterion \cite{Gibson2006HypothesisSF}.
The lattices are generated using a bigram LM.
A small constant learning rate with value 1e-5 is used.
At the final output layer, we use CE loss smoothing with a factor of 0.1.
The sMBR loss scale is set to 0.9.

\subsection{Language Models}
In recognition, we use 4-gram count-based language
model (LM) and LSTM LM as first pass decoding \cite{beck2019lstm}.
The LSTM LM has 51.3 perplexity (PPL) on Hub5’00 dataset.
We use a transformer (Trafo) LM for lattice rescoring having PPL
of 48.1 on Hub5’00 dataset.

\section{Experimental Results}
\label{sec:exp-results}

\subsection{Improved Baseline}
\label{sec:improved-baseline}

Our conformer acoustic model training recipe is built on top of our previous
work \cite{zeineldeen2022:hybrid-conformer}.
The improved results are reported in \Cref{tab:improved-baseline}.
% LN vs BN
We replace BN with LN for all convolution modules.
This also makes the model fits better for streaming mode.
This leads to $3\%$ relative improvement in terms of WER on Hub5'00.
% Adding another conv module
Moreover, we study the effect of adding another convolution module in each
conformer block following the Macaron style as suggested in
\cite{gulati2020conformer}.
This improves the system by $4\%$ relative in terms of WER on Hub5'00.
% reducing dimensions
In addition, we reduce the number of parameters by $40\%$ relative which
makes training more efficient and requires less memory usage.
To achieve that, we use an attention dimension of value 384 with 6 attention
heads.
The dimension of the feed-forward module is 1536.
We also remove LongSkip \cite{zeineldeen2022:hybrid-conformer}.
The total number of parameters is then only 58M .
The initial learning value is set to 1e-5.
We apply linear warmup over 5 epochs with a peak learning rate of 8e-4.
\Cref{tab:time-speed} shows that the improved baseline with reduced dimensions
is $27\%$ relatively faster in terms of training speed.

\begin{table}[t]
\centering
\caption{WERs [\%] of improved baseline. 4-gram LM is used for recognition.}
\label{tab:improved-baseline}
  \begin{tabular}{|l|c|c|c|c|}
    \hline
    \multirow{3}{*}{Model} & \multirow{3}{*}{\shortstack{Params. \\ $[\text{M}]$}}
    & \multicolumn{3}{c|}{WER [\%]} \\
    & & \multicolumn{3}{c|}{Hub5'00} \\ \cline{3-5}
    & & SWB & CH & Total \\ \hline
    Baseline & \multirow{2}{*}{88} & 8.0 & 15.4 & 11.7 \\ \cline{1-1} \cline{3-5}
    \hspace{1mm}+LN instead of BN & & 7.7 & 15.1 & 11.4 \\ \hline
    \hspace{3mm}+Two Conv modules & 98 & 7.3 & 14.7 & 11.0 \\ \hline
    \hspace{5mm}+Reduced dim. & 58 & \textbf{7.1} & \textbf{14.3} & \textbf{10.7} \\ \hline
  \end{tabular}
\end{table}

\begin{table}[t]
\centering
\caption{Comparison of training speed between the baseline conformer AM and
the reduced dimension one.
Training time is reported in hours for 1 epoch using
single GeForce GTX 1080 Ti GPU.
}
\label{tab:time-speed}
  \begin{tabular}{|l|c|c|}
    \hline
    Model & \shortstack{Train time [h]} \\ \hline
    Baseline & 5.5 \\ \hline
    \hspace{1mm}+Reduced dim. & 4.0 \\
    \hline
  \end{tabular}
\end{table}

\subsection{Speaker Adaptation Results}
\label{sec:sat-results}
% Observations:
% - weighted-simple-add is the best. mainly helps on CH
% - integrating at block 1 is the best
% - we integrate to the MHSA module

% setup
We use the best conformer AM as described in \Cref{sec:improved-baseline}
for speaker adaptive training (SAT).
We apply SAT by integrating i-vectors into the network.
We load the pretrained model and continue training with i-vectors while
adapting SpecAugment and applying learning rate reset to escape from a
local optimum as described in \cite{zhou2020rwth}.
We also do not apply any learning rate warmup because we start from a
pretrained model.

% method comparisons
In \Cref{tab:sat-methods}, we compare different methods to integrate i-vectors
into the conformer AM which are described in \Cref{sec:sat-methods}.
First, we can observe that we get more improvement on CH part compared to SWB
part since CH part is much noisier and contains more unseen speakers.
We can observe that {\weightedSimpleAdd} gives the larger improvement.
Using this method, we achieve $3.5\%$ relative improvement in terms of WER
on CH part of Hub5'00 and $3\%$ relative improvement overall.
On SWB part, all methods give a similar improvement which is only $0.1\%$
absolute.

% which block
For all SAT methods, we integrate the i-vectors into the first conformer block.
In \Cref{tab:pos-blocks}, we compare integrating i-vectors at different
blocks of the conformer where Block 0 refers to the front-end VGG network.
We can observe that integrating i-vectors at the first conformer block
gives the best improvement and the performance degrades as we use deeper blocks.
We can also observe that integrating at many blocks at the same time does not
help.
These results are done using the best method {\weightedSimpleAdd}.

% which module
Furthermore, we integrate i-vectors into the input of the MHSA module of
the first conformer block.
This means $z_t$ in the equations of \Cref{sec:sat-methods} represents
the input to the MHSA module.
We decide to use this module since it captures global context which can
potentially carry more speaker information.
To measure this experimentally, we report in \Cref{tab:module-spk-err} the
speaker error rates when adding a speaker identification loss on top of each
module in the first conformer block.
We can observe that the largest relative improvement is achieved at MHSA module.
The settings for speaker identification experiments are described in
\Cref{sec:spk-err-analysis}.

% some interpretation
% We can conclude that shifting the features is more important than scaling them.
% Same obervation is made in \cite{Rownicka2019}.

\begin{table}[t]
  \centering
  \caption{WERs [\%] of speaker adaptive training by using i-vectors with
  different integration methods. 4-gram LM is used for recognition.}
  \label{tab:sat-methods}
  \begin{tabular}{|l|c|c|c|}
  \hline
  \multirow{3}{*}{Integration method} &
  \multicolumn{3}{c|} {WER [\%]} \\ \cline{2-4}
  & \multicolumn{3}{c|}{Hub5'00} \\ \cline{2-4}
  & SWB & CH & Total \\ \hline
  - & 7.1  & 14.3 & 10.7 \\ \hline
  {\concat} & \textbf{7.0}  & 14.1 & 10.6 \\ \hline
  {\simpleAdd} & \textbf{7.0} & 14.0 & 10.5 \\ \hline
  {\complexAdd} & \textbf{7.0} & 14.0 & 10.5 \\ \hline
  {\gatedAdd} & \textbf{7.0} & 14.2 & 10.6  \\ \hline
  {\weightedSimpleAdd} & \textbf{7.0} & \textbf{13.8} & \textbf{10.4} \\ \hline
  \end{tabular}
\end{table}

\begin{table}[t]
  \centering
  \caption{WERs [\%] of speaker adaptive training by integrating i-vectors
  with {\weightedSimpleAdd} method at different conformer blocks.
  "(*)" means using multiple blocks. 4-gram LM is used for recognition.
  }
  \label{tab:pos-blocks}
  \begin{tabular}{|l|c|c|c|}
  \hline
  \multirow{3}{*}{Block} &
  \multicolumn{3}{c|} {WER [\%]} \\ \cline{2-4}
  & \multicolumn{3}{c|}{Hub5'00} \\ \cline{2-4}
  & SWB & CH & Total \\ \hline
  0 & 7.1 & 14.2 & 10.7 \\ \hline
  1 & \textbf{7.0} & \textbf{13.8} & \textbf{10.4} \\ \hline
  2 & 7.1 & 14.1 & 10.6 \\ \hline
  3 & 7.2 & 14.2 & 10.7 \\ \hline
  (1,2) & 7.1 & 14.0 & 10.5 \\ \hline
  (1,2,3) & 7.2 & 14.3 & 10.7 \\ \hline
  \end{tabular}
\end{table}

\begin{table}[t]
\centering
\caption{Speaker identification error rates using the output of different
modules in the first conformer block.}
\label{tab:module-spk-err}
  \begin{tabular}{|l|c|c|}
    \hline
    Module & Error [\%]\\ \hline
    %VGG block & 33.7 \\ \hline
    \nth{1} FFN & 31.7 \\ \hline
    \nth{1} Conv & 28.6 \\ \hline
    MHSA & 20.9 \\ \hline
    \nth{2} Conv & 19.6 \\ \hline
    \nth{2} FFN & 18.9 \\ \hline
  \end{tabular}
\end{table}

\subsection{Comparison between i-vector and x-vector}

The x-vectors are generated following the TDNN architecture in
\cite{snyder2018xvector}.
The training data is split into train and dev sets.
We use attentive pooling \cite{koji2018attentive} instead of statistical
pooling which improves the speaker identification accuracy from $87.5\%$ to
$93.2\%$ on the dev set.
The comparison between using i-vectors and x-vectors for speaker adaptation
using {\weightedSimpleAdd} method is reported in
\Cref{tab:embedding-comparision}.
Using x-vectors did not improve the performance but instead, lead to minor
degradation on CH part.
On possible explanation is that x-vectors mainly focus on the speaker
characteristics whereas i-vectors are designed to capture both speaker
and channel characteristics which might be important in this case
\cite{Rownicka2019}.
Further investigations are needed for experiments with x-vectors.

\begin{table}[t]
  \centering
  \caption{WERs [\%] of comparison between i-vectors and x-vectors for
  speaker adaptation using {\weightedSimpleAdd} integration method.
  4-gram LM is used for recognition.}
  \label{tab:embedding-comparision}
  \begin{tabular}{|c|c|c|c|}
  \hline
  \multirow{3}{*}{Embedding}  & \multicolumn{3}{c|}{WER [\%]} \\ \cline{2-4}
  & \multicolumn{3}{c|}{Hub5'00} \\ \cline{2-4}
  & SWB & CH & Total \\ \hline
  i-vector  & \textbf{7.0} & \textbf{13.8} & \textbf{10.4} \\ \hline
  x-vector  & 7.0 & 14.4 & 10.7 \\ \hline
  \end{tabular}
\end{table}

\subsection{Overall Results}

We summarize our results in \Cref{tab:overall_results} and compare with
different modeling approaches and architectures from the literature.
Compared to our previous work \cite{zeineldeen2022:hybrid-conformer}, this work
achieves $11.0\%$ and $4.1\%$ relative improvement with transformer LM on
Hub5'00 and Hub5'01 datasets respectively.
Moreover, our model outperforms a well-optimized BLSTM attention system
\cite{tuske2020single} by $3\%$ and $5\%$ with LSTM LM on Hub5'00 and Hub5'01
datasets respectively.
In addition, our model outperforms the TDNN baseline by $9\%$ relative
on Hub5'00 dataset.
Our best conformer AM is still behind the state-of-the-art results but not
with a big margin.
We argue that our baseline can be further improved by applying speed
perturbation \cite{tuske2020single} and also longer training.
Surprisingly, sequence training only give minor improvement which requires further
investigation.

\begin{table}[t]
  \caption{Overall WER [\%] comparison with literature.}
  \label{tab:overall_results}
  \centering
  \resizebox{\columnwidth}{!} {
  \setlength\tabcolsep{1pt}
  \begin{tabular}{|@{\hskip1pt}c@{\hskip1pt}
    |@{\hskip2pt}c@{\hskip2pt}|@{\hskip1pt}c@{\hskip1pt}|@{\hskip1pt}c@{\hskip1pt}|
    @{\hskip1pt}c@{\hskip1pt}|@{\hskip1pt}c@{\hskip1pt}
    |@{\hskip1pt}c@{\hskip1pt}|@{\hskip1pt}c@{\hskip1pt}|
    @{\hskip1pt}c@{\hskip1pt}|@{\hskip1pt}c@{\hskip1pt}|}
    \hline
    \multirow{3}{*}{Work} & \multirow{3}{*}{Approach}
    & \multirow{3}{*}{AM} & \multirow{3}{*}{LM} &  \multirow{3}{*}{\shortstack{seq.\\train}} &
    \multirow{3}{*}{\shortstack{ivec.}} &
    \multicolumn{4}{c|}{WER [\%]} \\
    \cline{7-10}
    & & & & & &   \multicolumn{3}{c|}{Hub5'00}  & \multirow{2}{*}{\shortstack{Hub\\5'01}} \\ \cline{7-9}
    & & & & & & SWB & CH & Total & \\ \hline

    % TDNN
    \cite{hu2021bayesian} & Hybrid & TDNN & RNN & yes & yes &
    7.2 & 13.6 & 10.4 & - \\ \hline

    % old baseline
    \multirow{2}{*}{\cite{zeineldeen2022:hybrid-conformer}} & \multirow{2}{*}{Hybrid}
    & \multirow{2}{*}{Conf.} & LSTM & \multirow{2}{*}{yes} &
    \multirow{2}{*}{no} & 6.9 & 14.5 & 10.7 & 10.1 \\ \cline{4-4} \cline{7-10}
    & & &  Trafo & & & 6.6 & 14.1 & 10.3 & \phantom{0}9.7 \\ \hline

    % Saon's RNN-T paper
    \cite{Saon2021RNN-T} & RNN-T & LSTM & LSTM & no & yes & 6.3
    & 13.1 & \phantom{0}9.7 & 10.1 \\ \hline

    % Zoltan's LSTM
    \cite{tuske2020single}
    & LAS & LSTM & LSTM & no & no & 6.5 & 13.0 & \phantom{0}9.8 & 10.1 \\ \hline

    % Zoltan's conformer
    \multirow{3}{*}{\cite{Zoltan2021conformer}}
    & \multirow{3}{*}{LAS} & \multirow{3}{*}{Conf.} & - &
    \multirow{3}{*}{no} & \multirow{3}{*}{yes} & 6.7 & 13.0 & \phantom{0}9.9 & 10.1 \\
    \cline{4-4} \cline{7-10}
    &   &  & LSTM & & & 5.7 & 11.4 & \phantom{0}8.6 & \phantom{0}8.5 \\ \cline{4-4} \cline{7-10}
    &   &  & Trafo & & & 5.5 & 11.2 & \phantom{0}8.4 & \phantom{0}8.5 \\
    \hline \hline

    % ours
    \multirow{5}{*}{\shortstack{this\\work}} & \multirow{5}{*}{Hybrid} & \multirow{5}{*}{Conf.}
    & \multirow{3}{*}{4-gram} & \multirow{2}{*}{no} & no  & 7.1 & 14.3 & 10.7 & 11.0 \\ \cline{6-6} \cline{7-10}
    &  & & & & \multirow{4}{*}{yes} & 7.0 & 13.8 & 10.4 & 10.5  \\  \cline{5-5} \cline{7-10}
    &  & & & \multirow{3}{*}{yes} &  & 7.1 & 13.5 & 10.3 & 10.4  \\
     \cline{4-4} \cline{7-10}
    &  & & LSTM & & & 6.5 & 12.5 & \phantom{0}9.5 & \phantom{0}9.6 \\
     \cline{4-4} \cline{7-10}
    &  & & Trafo & &  & \textbf{6.3} & \textbf{12.1} & \textbf{\phantom{0}9.2} & \textbf{\phantom{0}9.3} \\ \hline
  \end{tabular}
}
\end{table}

\section{Speaker Identification Error Analysis}
\label{sec:spk-err-analysis}

\begin{figure}[t]
\begin{tikzpicture}
\begin{axis}[
    width=0.45\textwidth,
    height=0.3\textwidth,
    ylabel={Speaker Identification Error [\%]},
    xlabel={Layer Depth},
    xmin=0, xmax=1.0,
    ymin=0, ymax=90.0,
    legend pos=north west,
    ymajorgrids=true,
    grid style=dashed,
]

\addplot[color=blue,mark=triangle,]
    coordinates {
    (0.0, 34)
    (0.083, 19)
    (0.17, 13)
    (0.25, 14)
    (0.33, 21)
    (0.42, 22)
    (0.50, 25)
    (0.58, 28)
    (0.67, 37)
    (0.75, 41)
    (0.83, 43)
    (0.92, 47)
    (1.0,  55)
    };

\addplot[color=green,mark=square,]
    coordinates {
    (0.0, 47)
    (0.2, 47)
    (0.4, 43)
    (0.6, 43)
    (0.8, 57)
    (1.0, 83)
    };
\legend{Conformer, BLSTM}
\end{axis}
\end{tikzpicture}
\caption{Speaker identification error rates using BLSTM AM
and conformer AM. Layer depth refers to the layer index divided by total
number of layers.}
%Layer 0 refers to the front-end VGG network of conformer AM.}
\label{fig:spk-err}
\vspace{-5mm}
\end{figure}
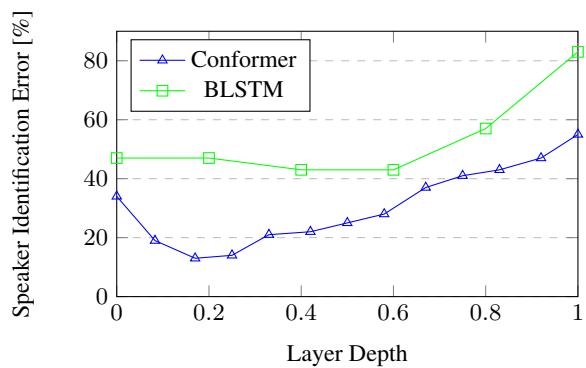

It has been observed empirically that the improvement by speaker adaptation
is not significant when using a conformer-based AM \cite{Zoltan2021conformer}.
However, BLSTM-based AMs benefit much more from speaker adaptation
\cite{kitza2019cumulative}.
We argue that the reason behind this is that the conformer model is able to
better learn speaker information in the lower layers which make it already
robust to different speakers.
To verify this, we conduct the following experiment using the Switchboard
300h training dataset.
We added a speaker identification loss on top of the output of each
layer for both the conformer model and BLSTM model.
The BLSTM model consists of 6 layers following
\cite{zeineldeen2022:hybrid-conformer,kitza2019cumulative}.
Then, we freeze all the AM parameters except the speaker identification loss
parameters.
We use an attention mechanism as \cite{Siddharth2020} to get a soft
speaker representation from the last output layer of the AM.
This can also learn to skip unnecessary information such as silence.
The total number of speakers is 520.
We split the training dataset into train and dev parts.
The speaker identification error rates on the dev set are reported in
\Cref{fig:spk-err}.
We can observe that the conformer model has a much better speaker
identification accuracy using the lower layers compared to the BLSTM model.
It was able to achieve using the second block a speaker identification
error rate of $13\%$ while the lowest error rate that the BLSTM achieves
is $43\%$.
These figures also show that lower layers are more important for learning
speaker representations.
For the conformer, we also observe that having the speaker loss on top of the
MHSA module is enough to achieve such a low speaker error rate.
This also explains why the proposed method to integrate the i-vectors into
the MHSA module works the best for us.
Moreover, the speaker error starts to increase for higher layers which is
most probably because the model starts to filter out unnecessary speaker
information to learn high-level features.

\section{Conclusions \& Future Work}
% Future work:
% - model-space methods
% - try our recipe on more noisy data such as Chime4
% Findings:
% - possible to build smaller and efficient model with good performance
% - propose a better way for SAT that improves on CH part
% - i-vector works better than x-vector
% - conformer seems to work well for speaker identification tasks

In this work, we improved our training recipe for conformer-based
hybrid Neural Network-Hidden Markov Model.
We achieved $11\%$ relative improvement in terms of word-error-rate (WER) on
the Switchboard 300h Hub5'00 dataset.
We also developed an efficient training recipe with $34\%$ relative reduction
in total number of parameters.
Moreover, we investigated different ways to apply speaker adaptive training
(SAT) for conformer-based acoustic models.
We propose a method, called {\weightedSimpleAdd} to add weighted speaker
information vectors to the input of multi-head self-attention module of
the conformer AM.
Using this method for SAT, we achieve $3.5\%$ and $4.5\%$ relative
improvement in terms of WER on the CallHome part of Hub5'00 and Hub5'01
respectively.
We also perform some analysis and showed that the conformer-based AM is already
good enough for speaker identification task.
For future work, we plan to apply our recipe to a noisier dataset such
as CHiME-4 \cite{Yang2022conformer} as well as applying model-space SAT
approaches \cite{Markus2018AT}.

\section{Acknowledgements}
This work was partially supported by the project HYKIST funded by the
German Federal Ministry of Health on the basis of a decision of the
German Federal Parliament (Bundestag) under funding ID ZMVI1-2520DAT04A.
We thank Wilfried Michel, Zoltán Tüske, and Wei Zhou for useful discussions.
We also thank Alexander Gerstenberger for performing lattice rescoring with
Transformer.

\bibliographystyle{IEEEtran}
\bibliography{refs}

% Generated by IEEEtran.bst, version: 1.13 (2008/09/30)
\begin{thebibliography}{10}
\providecommand{\url}[1]{#1}
\csname url@samestyle\endcsname
\providecommand{\newblock}{\relax}
\providecommand{\bibinfo}[2]{#2}
\providecommand{\BIBentrySTDinterwordspacing}{\spaceskip=0pt\relax}
\providecommand{\BIBentryALTinterwordstretchfactor}{4}
\providecommand{\BIBentryALTinterwordspacing}{\spaceskip=\fontdimen2\font plus
\BIBentryALTinterwordstretchfactor\fontdimen3\font minus
  \fontdimen4\font\relax}
\providecommand{\BIBforeignlanguage}[2]{{%
\expandafter\ifx\csname l@#1\endcsname\relax
\typeout{** WARNING: IEEEtran.bst: No hyphenation pattern has been}%
\typeout{** loaded for the language `#1'. Using the pattern for}%
\typeout{** the default language instead.}%
\else
\language=\csname l@#1\endcsname
\fi
#2}}
\providecommand{\BIBdecl}{\relax}
\BIBdecl

\bibitem{zeineldeen2022:hybrid-conformer}
M.~Zeineldeen, J.~Xu, C.~L\"uscher, W.~Michel, A.~Gerstenberger, R.~Schl\"uter,
  and H.~Ney, ``{Conformer-based Hybrid ASR System for Switchboard Dataset},''
  in \emph{ICASSP}, Singapore, May 2022, pp. 7437--7441.

\bibitem{zhou2020rwth}
W.~Zhou, W.~Michel, K.~Irie, M.~Kitza, R.~Schlüter, and H.~Ney, ``{The RWTH
  ASR System for TED-LIUM Release 2: Improving Hybrid HMM with SpecAugment},''
  in \emph{ICASSP}, Barcelona, Spain, May 2020, pp. 7839--7843.

\bibitem{Lscher2019RWTHAS}
C.~Lüscher, E.~Beck, K.~Irie, M.~Kitza, W.~Michel, A.~Zeyer, R.~Schlüter, and
  H.~Ney, ``{RWTH ASR Systems for LibriSpeech: Hybrid vs Attention},'' in
  \emph{INTERSPEECH}, Graz, Austria, Sep. 2019, pp. 231--235.

\bibitem{hochreiter1997long}
S.~Hochreiter and J.~Schmidhuber, ``{Long Short-term Memory},'' \emph{{Neural
  Computation}}, vol.~9, no.~8, pp. 1735--1780, 1997.

\bibitem{abdel2014convolutional}
O.~Abdel-Hamid, A.~Mohamed, H.~Jiang, L.~Deng, G.~Penn, and D.~Yu,
  ``{Convolutional Neural Networks for Speech Recognition},'' \emph{{IEEE/ACM
  Transactions on Audio, Speech, and Language Processing}}, vol.~22, no.~10,
  pp. 1533--1545, 2014.

\bibitem{peddinti2015time}
V.~Peddinti, D.~Povey, and S.~Khudanpur, ``{A Time Delay Neural Network
  Architecture for Efficient Modeling of Long Temporal Contexts},'' in
  \emph{INTERSPEECH}, Dresden, Germany, Sep. 2015, pp. 3214--3218.

\bibitem{Sperber2018SelfAttentionalAM}
M.~Sperber, J.~Niehues, G.~Neubig, S.~St{\"{u}}ker, and A.~Waibel,
  ``{Self-Attentional Acoustic Models},'' in \emph{INTERSPEECH}, Hyderabad,
  India, Sep. 2018, pp. 3723--3727.

\bibitem{gulati2020conformer}
A.~Gulati, J.~Qin, C.~Chiu, N.~Parmar, Y.~Zhang, J.~Yu, W.~Han, S.~Wang,
  Z.~Zhang, Y.~Wu, and R.~Pang, ``{Conformer: Convolution-augmented Transformer
  for Speech Recognition},'' in \emph{INTERSPEECH}, Shanghai, China, Oct. 2020,
  pp. 5036--5040.

\bibitem{Zoltan2021conformer}
Z.~T{\"{u}}ske, G.~Saon, and B.~Kingsbury, ``{On the Limit of English
  Conversational Speech Recognition},'' in \emph{INTERSPEECH}, Brno, Czechia,
  Sep. 2021, pp. 2062--2066.

\bibitem{Saon2013SAT}
G.~Saon, H.~Soltau, D.~Nahamoo, and M.~Picheny, ``{Speaker Adaptation of Neural
  Network Acoustic Models Using I-Vectors},'' in \emph{ASRU}, Olomouc, Czech
  Republic, Dec. 2013, pp. 55--59.

\bibitem{mllr}
\BIBentryALTinterwordspacing
C.~Leggetter and P.~Woodland, ``{Maximum Likelihood Linear Regression for
  Speaker Adaptation of Continuous Density Hidden Markov Models},''
  \emph{Computer Speech \& Language}, vol.~9, no.~2, pp. 171--185, 1995.
  [Online]. Available:
  \url{https://www.sciencedirect.com/science/article/pii/S0885230885700101}
\BIBentrySTDinterwordspacing

\bibitem{Dehak2011Ivector}
N.~Dehak, P.~J. Kenny, R.~Dehak, P.~Dumouchel, and P.~Ouellet, ``{Front-End
  Factor Analysis for Speaker Verification},'' \emph{IEEE Transactions on
  Audio, Speech, and Language Processing}, vol.~19, no.~4, pp. 788--798, 2011.

\bibitem{snyder2018xvector}
D.~Snyder, D.~Garcia-Romero, G.~Sell, D.~Povey, and S.~Khudanpur, ``{X-Vectors:
  Robust {DNN} Embeddings for Speaker Recognition},'' in \emph{ICASSP},
  Calgary, Alberta, Canada, 2018, pp. 5329--5333.

\bibitem{seide2011feature}
F.~Seide, G.~Li, X.~Chen, and D.~Yu, ``{Feature Engineering in
  Context-dependent Deep Neural Networks for Conversational Speech
  Transcription},'' in \emph{ASRU}, Waikoloa, HI, USA, 2011, pp. 24--29.

\bibitem{gemello2006adaptation}
R.~Gemello, F.~Mana, S.~Scanzio, P.~Laface, and R.~De~Mori, ``{Adaptation of
  Hybrid ANN/HMM Models using Linear Hidden Transformations and Conservative
  Training},'' in \emph{ICASSP}, vol.~1.\hskip 1em plus 0.5em minus 0.4em\relax
  Toubouse, France: IEEE, May 2006, pp. I--I.

\bibitem{li2010comparison}
B.~Li and K.~C. Sim, ``{Comparison of Discriminative Input and Output
  Transformations for Speaker Adaptation in the Hybrid NN/HMM Systems},'' in
  \emph{Eleventh Annual Conference of the International Speech Communication
  Association}, Makuhari, Chiba, Japan, 2010.

\bibitem{Zhu2019multiaccent}
H.~Zhu, L.~Wang, P.~Zhang, and Y.~Yan, ``{Multi-accent Adaptation Based on Gate
  Mechanism},'' in \emph{INTERSPEECH}, Graz, Austria, Sep. 2019, pp. 744--748.

\bibitem{gong2021multiaccent}
X.~Gong, Y.~Lu, Z.~Zhou, and Y.~Qian, ``{Layer-wise Fast Adaptation for
  End-to-End Multi-accent Speech Recognition},'' in \emph{INTERSPEECH}, Brno,
  Czechia, 2021, pp. 1274--1278.

\bibitem{deng2021improving}
K.~Deng, S.~Cao, and L.~Ma, ``{Improving Accent Identification and Accented
  Speech Recognition under a Framework of Self-supervised Learning},'' in
  \emph{INTERSPEECH}, Brno, Czechia, Sep. 2021, pp. 1504--1508.

\bibitem{turan2020multiaccent}
M.~A.~T. Turan, E.~Vincent, and D.~Jouvet, ``{Achieving Multi-accent {ASR} via
  Unsupervised Acoustic Model Adaptation},'' in \emph{INTERSPEECH}, Shanghai,
  China, Oct. 2020, pp. 1286--1290.

\bibitem{Ioffe2015bn}
S.~Ioffe and C.~Szegedy, ``{Batch Normalization: Accelerating Deep Network
  Training by Reducing Internal Covariate Shift},'' in \emph{{ICML}}, vol.~37,
  Lille, France, 2015, pp. 448--456.

\bibitem{Lei2016ln}
L.~J. Ba, J.~R. Kiros, and G.~E. Hinton, ``{Layer Normalization},''
  \emph{CoRR}, vol. abs/1607.06450, 2016.

\bibitem{Peddinti2015tdnn}
V.~Peddinti, G.~Chen, V.~Manohar, T.~Ko, D.~Povey, and S.~Khudanpur, ``{{JHU}
  ASpIRE System: Robust {LVCSR} with TDNNS, IVector Adaptation and
  {RNN-LMS}},'' in \emph{{ASRU}}.\hskip 1em plus 0.5em minus 0.4em\relax
  Scottsdale, AZ, USA: {IEEE}, Dec. 2015, pp. 539--546.

\bibitem{kitza2019cumulative}
M.~Kitza, P.~Golik, R.~Schlüter, and H.~Ney, ``{Cumulative Adaptation for
  BLSTM Acoustic Models},'' in \emph{INTERSPEECH}, Graz, Austria, Sep. 2019,
  pp. 754--758.

\bibitem{yoo2019film}
S.~Yoo, I.~Song, and Y.~Bengio, ``{A Highly Adaptive Acoustic Model for
  Accurate Multi-dialect Speech Recognition},'' in \emph{ICASSP}, Brighton, UK,
  May 2019, pp. 5716--5720.

\bibitem{Rownicka2019}
J.~Rownicka, P.~Bell, and S.~Renals, ``{Embeddings for {DNN} Speaker Adaptive
  Training},'' in \emph{{ASRU}}.\hskip 1em plus 0.5em minus 0.4em\relax
  Singapore: {IEEE}, Dec. 2019, pp. 479--486.

\bibitem{Godfrey1992SWB}
J.~Godfrey, E.~Holliman, and J.~McDaniel, ``{Switchboard: Telephone Speech
  Corpus for Research and Development},'' in \emph{ICASSP}, vol.~1, San
  Francisco,USA, Mar. 1992, pp. 517--520.

\bibitem{Zeyer_2018}
A.~Zeyer, T.~Alkhouli, and H.~Ney, ``{RETURNN as a Generic Flexible Neural
  Toolkit with Application to Translation and Speech Recognition},'' in
  \emph{Annual Meeting of the Assoc. for Computational Linguistics}, Melbourne,
  Australia, Jul. 2018.

\bibitem{Wiesler2014RASR}
S.~Wiesler, A.~Richard, P.~Golik, R.~Schlüter, and H.~Ney, ``{RASR/NN: The
  RWTH Neural Network Toolkit for Speech Recognition},'' in \emph{ICASSP},
  Florence, Italy, May 2014, pp. 3313--3317.

\bibitem{long2015fully}
J.~Long, E.~Shelhamer, and T.~Darrell, ``{Fully Convolutional Networks for
  Semantic Segmentation},'' in \emph{CVPR}, Boston, USA, Jun. 2015, pp.
  3431--3440.

\bibitem{schluter2007Gammatone}
R.~Schluter, I.~Bezrukov, H.~Wagner, and H.~Ney, ``{Gammatone Features and
  Feature Combination for Large Vocabulary Speech Recognition},'' in
  \emph{ICASSP}, Honolulu, USA, Apr. 2007, pp. 649--652.

\bibitem{Rouhe2020}
A.~Rouhe, T.~Kaseva, and M.~Kurimo, ``{Speaker-aware Training of
  Attention-based End-to-end Speech Recognition Using Neural Speaker
  Embeddings},'' in \emph{ICASSP}, Barcelona, Spain, May 2020, pp. 7064--7068.

\bibitem{Gibson2006HypothesisSF}
M.~Gibson and T.~Hain, ``{Hypothesis Spaces for Minimum Bayes Risk Training in
  Large Vocabulary Speech Recognition},'' in \emph{{INTERSPEECH}}, Pittsburgh,
  USA, Sep. 2006.

\bibitem{beck2019lstm}
\BIBentryALTinterwordspacing
E.~Beck, W.~Zhou, R.~Schlüter, and H.~Ney, ``{{LSTM} Language Models for
  {LVCSR} in First-pass Decoding and Lattice-Rescoring},'' \emph{CoRR}, vol.
  abs/1907.01030, 2019. [Online]. Available:
  \url{http://arxiv.org/abs/1907.01030}
\BIBentrySTDinterwordspacing

\bibitem{koji2018attentive}
K.~Okabe, T.~Koshinaka, and K.~Shinoda, ``{Attentive Statistics Pooling for
  Deep Speaker Embedding},'' in \emph{INTERSPEECH}, Hyderabad, India, Sep.
  2018, pp. 2252--2256.

\bibitem{tuske2020single}
Z.~Tüske, G.~Saon, K.~Audhkhasi, and B.~Kingsbury, ``{Single Headed Attention
  Based Sequence-to-sequence Model for State-of-the-Art Results on
  Switchboard},'' in \emph{INTERSPEECH}, Shanghai, China, Sep. 2020, pp.
  551--555.

\bibitem{hu2021bayesian}
S.~Hu, X.~Xie, S.~Liu, J.~Yu, Z.~Ye, M.~Geng, X.~Liu, and H.~Meng, ``{Bayesian
  Learning of LF-MMI Trained Time Delay Neural Networks for Speech
  Recognition},'' \emph{IEEE/ACM Transactions on Audio, Speech, and Language
  Processing}, vol.~29, pp. 1514--1529, 2021.

\bibitem{Saon2021RNN-T}
G.~Saon, Z.~T{\"{u}}ske, D.~Bola{\~{n}}os, and B.~Kingsbury, ``{Advancing {RNN}
  Transducer Technology for Speech Recognition},'' in \emph{{ICASSP}}.\hskip
  1em plus 0.5em minus 0.4em\relax Toronto, Canada: {IEEE}, 2021, pp.
  5654--5658.

\bibitem{Siddharth2020}
S.~Sigtia, E.~Marchi, S.~Kajarekar, D.~Naik, and J.~Bridle, ``{Multi-Task
  Learning for Speaker Verification and Voice Trigger Detection},'' in
  \emph{{ICASSP}}.\hskip 1em plus 0.5em minus 0.4em\relax Barcelona, Spain:
  {IEEE}, May 2020, pp. 6844--6848.

\bibitem{Yang2022conformer}
\BIBentryALTinterwordspacing
Y.~Yang, P.~Wang, and D.~Wang, ``{A Conformer Based Acoustic Model for Robust
  Automatic Speech Recognition},'' \emph{CoRR}, vol. abs/2203.00725, 2022.
  [Online]. Available: \url{https://doi.org/10.48550/arXiv.2203.00725}
\BIBentrySTDinterwordspacing

\bibitem{Markus2018AT}
M.~Kitza, R.~Schl{\"{u}}ter, and H.~Ney, ``{Comparison of BLSTM-layer-specific
  Affine Transformations for Speaker Adaptation},'' in \emph{INTERSPEECH},
  Hyderabad, India, Sep. 2018, pp. 877--881.

\end{thebibliography}

\end{document}